\documentclass[conference]{IEEEtran}
\IEEEoverridecommandlockouts
\usepackage{cite}
\usepackage{url}
\usepackage{booktabs}
\usepackage{amsmath,amssymb,amsfonts}
\usepackage{algorithmic}
\usepackage{graphicx}
\usepackage{lipsum}
\usepackage[T1]{fontenc}
\usepackage{textcomp}
\usepackage{tikz}
\usepackage{atbegshi}
\usepackage{xcolor}
\usepackage{csquotes}
\usepackage{comment}
\usepackage{hyperref}
 \hypersetup{
colorlinks=true,
linkcolor=blue,
filecolor=blue,
citecolor = blue,
urlcolor=blue,
}
\def\BibTeX{{\rm B\kern-.05em{\sc i\kern-.025em b}\kern-.08em
    T\kern-.1667em\lower.7ex\hbox{E}\kern-.125emX}}
\begin{document}

\title{Smart IoT Security: Lightweight Machine Learning Techniques for Multi-Class Attack Detection in IoT Networks}

\author{
\IEEEauthorblockN{Shahran Rahman Alve}
\IEEEauthorblockA{\textit{Dept. of ECE} \\
\textit{North South University}\\
Dhaka-1229, Bangladesh \\
shahran.alve@northsouth.edu}
\and
\IEEEauthorblockN{Muhammad Zawad Mahmud}
\IEEEauthorblockA{\textit{Dept. of ECE} \\
\textit{North South University}\\
Dhaka-1229, Bangladesh \\
zawad.mahmud1@northsouth.edu}
\and
\IEEEauthorblockN{Samiha Islam}
\IEEEauthorblockA{\textit{Dept. of ECE} \\ 
\textit{North South University}\\
Dhaka-1229, Bangladesh \\
samiha.islam2@northsouth.edu}
\and 
\IEEEauthorblockN{Md. Asaduzzaman Chowdhury}
\IEEEauthorblockA{\textit{Dept. of ECE} \\
\textit{North South University}\\
Dhaka-1229, Bangladesh \\
asaduzzaman.chowdhury@northsouth.edu}
\and 
\IEEEauthorblockN{Jahirul Islam}
\IEEEauthorblockA{\textit{Dept. of ECE} \\
\textit{North South University}\\
Dhaka-1229, Bangladesh \\
jahirul.islam04@northsouth.edu}
}

\maketitle
\begin{tikzpicture}[remember picture, overlay]
    \node[anchor=north west, xshift=1.5cm, yshift=-0.5cm] at (current page.north west)
    {2025 International Conference on Quantum Photonics, Artificial Intelligence, and Networking (QPAIN)};
\end{tikzpicture}
\begin{tikzpicture}[remember picture, overlay]
    \node[anchor=north west, xshift=1.5cm, yshift=-1cm] at (current page.north west)
    {31 July - 2 August 2025, Rangpur, Bangladesh};
\end{tikzpicture}
\IEEEpubid{\begin{minipage}{\textwidth}\ \\[60pt]
979-8-3315-9694-1/25/\$31.00 ©2025 IEEE
\end{minipage}}

\begin{abstract}
The Internet of Things (IoT) is expanding at an accelerated pace, making it critical to have secure networks to mitigate a variety of cyber threats. This study addresses the limitation of multi-class attack detection of IoT devices and presents new machine learning-based lightweight ensemble methods that exploit its strong machine learning framework. We used a dataset entitled CICIoT 2023, which has a total of 34 different attack types categorized into 10 categories, and methodically assessed the performance of a substantial array of current machine learning techniques in our goal to identify the best-performing algorithmic choice for IoT application protection. In this work, we focus on ML classifier-based methods to address the biocharges presented by the difficult and heterogeneous properties of the attack vectors in IoT ecosystems. The best-performing method was the Decision Tree, achieving 99.56\% accuracy and 99.62\% F1, indicating this model is capable of detecting threats accurately and reliably. The Random Forest model also performed nearly as well, with an accuracy of 98.22\% and an F1 score of 98.24\%, indicating that ML methods excel in a scenario of high-dimensional data. These findings emphasize the promise of integrating ML classifiers into the protective defenses of IoT devices and provide motivations for pursuing subsequent studies towards scalable, keystroke-based attack detection frameworks. We think that our approach offers a new avenue for constructing complex machine learning algorithms for low-resource IoT devices that strike a balance between accuracy requirements and time efficiency. In summary, these contributions expand and enhance the knowledge of the current IoT security literature, establishing a solid baseline and framework for smart, adaptive security to be used in IoT environments.
\end{abstract}


\section{Introduction}
The growing Internet of Things (IoT) technology is evolving rapidly, revolutionizing industries by enabling secure communication and facilitating the smart interaction between devices and systems, significantly enhancing automation and data-driven decision-making processes. Though this expansion of Internet of Things devices offers us several advantages, it also endangers us with a variety of cyberattacks. The security and privacy threats of IoT devices are primarily caused by the low-budget implementations of IoT due to the limited access to computing resources in IoT devices, which results in different kinds of attacks such as Distributed Denial of Service (DDoS), brute-force attacks, and data tampering. As per a Kaspersky analysis, the IoT attacks increased to 1.51 billion within the first half of 2021, a 100\% increase compared with 639 million over a similar timeframe in 2020 \cite{kaspersky2021iot}. As IoT networks are growing in size and complexity, there is a need for efficient intrusion detection systems (IDS) to protect these interconnected systems against advanced persistent threats.As emphasized by Al-Garadi et al. \cite{algaradi2020survey}, advanced security solutions for IoT networks are being implemented using ML and DL. Augmented with these approaches, security systems transform from legacy ways to intelligent, data-intensive systems that can learn and detect emerging threats in real time. With it, attacks can finish in a more diverse and quick response than conventional and rigid security systems. However, traditional IDS tends to have a high false positive rate and security system confusion since it cannot effectively cope with a novel kind of attack. In addressing these limitations, approaches using machine learning have demonstrated the capabilities to improve the detection and classification of IoT attacks (Sarker et al. \cite{sarker2023iot}). In this paper, a lightweight classification-based ensemble machine learning technique is proposed to enhance the detection of multiple types of attacks in the IoT network. The study will involve developing an IDS that accurately recognizes both conventional and novel attack types by performing hyperparameter tuning and feature extraction. This novel approach provides a more efficient and adaptable solution for enhancing IDS predictive and responsive effectiveness, particularly in the context of IoT security. The primary accomplishments of this work are as follows:
\begin{itemize}
\item Utilizing the machine learning methods with hyperparameter tuning with the help of the research aims to extract essential features from a dataset containing both attacks and normal situations.
\item With a testing accuracy of 99.56\%, the proposed Decision Tree with hyperparameter tuning aims to create a dependable, economical, and rapid diagnostic solution characterized by high accuracy and robust validation measures.
\end{itemize}

\section{Related Work}
Nowadays, Many IoT devices are constructed based on the low-power lossy networks, e.g., Routing Protocol for Low-Power and Lossy Networks (RPL), as it creates a great opportunity for many types of security attacks because these devices have constraints on their computational resources. Due to such a limitation of these networks, the traditional IDS approach could not adequately oversee and secure the IoT device traffic. Al Sawafi et al. \cite{al_sawafi2023hybrid} discussed in detail the issue and suggested an IDS mechanism based on deep learning models, having both the supervised and semi-supervised classification algorithms to distinguish the category of network attacks regarding type of known and type of unseen attacks. They took sixteen researched attacks to save the data through packet sniffing; the results reveal that their deep learning-based model achieved 99\% accuracy and 98\% F1-score in identifying known attacks, which denotes deep learning plays a vital role in influencing IoT security positively.

Mahmud et al. \cite{10928532} utilized a large IoT dataset containing 820,834 data samples with 54 features to evaluate the performance of various ML classifiers for network intrusion detection, which makes its contribution much more significant. The baseline dataset containing equal instances of normal and attack cases formed the backbone of their analyses. Similarly, their findings showed that the best accuracy of the Random Forest Classifier was 99.39\%. 

Zakariah et al. \cite{zakariah2023machine} introduced a novel intrusion detection model to secure IoT networks by addressing the limitations of current firewalls and encryption techniques to mitigate emerging threats in networks. They applied CNN and LSTM networks with attention mechanisms and complemented them with adaptive synthetic sampling of imbalanced datasets. The hybrid model produced an accuracy of 89\%, and an F1 score of 90\%, compared with the MLP baseline with an accuracy of 87\% and an F1 score of 88\%.

Musleh et al. \cite{musleh2023intrusion} highlight that feature extraction crucially increases the efficiency of ML-based IDS for the IoT networks. In addition, the IEEE Dataport dataset is used in the paper to validate that by using the VGG-16 model combined with stacked ensemble classifiers, including a combining classifier based on KNN and sequential minimum optimization (SMO), one can reach an accuracy of 98.3\%, while maintaining perfect precision.

Chaganti et al. \cite{chaganti2023deep} present an enhanced LSTM networks-based deep learning approach for intrusion detection to protect SDNL enabled IoT networks, and this work boosts the detection and classification of attacks. It achieved a 97.7\% accuracy in classifying each multi-class attack when applied to two datasets that incorporate network threat techniques that include SDNIoT, where the LSTM model is able to avoid DDoS, surveillance, and other forms of network attacks.

Khan et al. \cite{khan2023hybrid} created a hybrid deep learning-based intrusion detection system that can detect intrusions of IoT systems through physical, network, and application levels using recurrent neural networks and gated recurrent units. An advanced deep learning model, optimized via the Adam and Adamax methods, was used to evaluate the performance of the proposed model over the ToN-IoT dataset. The evaluation indicates high accuracy of 97\% and an F1-score of 96\%, for the Adamax optimizer.

Rose et al. \cite{rose2021intrusion} proposed a dataset and model to investigate the efficiency of network characterization and machine learning to safeguard IoT devices against cyber-attacks. Empirical data indicate that the proposed anomaly detection system obtains 98.35\% accuracy and 98.35\% false-positive alarms.

Mohy-Eddine et al. \cite{mohy-eddine2022ensemble} introduced a new IDS model to a more secure Industrial Internet of Things (IIoT); They studied some of the vulnerabilities that are standard in the IIoT environments. This approach uses Isolation Forest (IF) and Pearson Correlation Coefficient (PCC) in the feature engineering and employs Random Forest classifiers to improve the detection performance. As highlighted above, the SNCF-PCCIF was a promising classifier on the UNSW-NB15-v2 datasets, equal to RF-PCCIF with 99.30\% accuracy.


With the purpose of addressing these problems, Potnurwar et al. \cite{potnurwar2023deep} emphasized a new model based on deep learning, hybrid feature selection, along with DFFNN and rule-based hybrid feature selection methods. The datasets used for the research include NSL-KDD and UNSW-NB15. For their best-performing test using F1 scores, the researchers were able to derive a true positive rate of a very high 96\% in attack and 98\% in normal.

\section{Methodology}
\subsection{Dataset}
This dataset used in this research is the latest CICIoT 2023 dataset by Neto et al. \cite{neto2023ciciot2023}. This dataset has more IoT networks and IoT device targets and intruders. This dataset consists of 34 types of attacks grouped by 10 categories: DDoS attacks, DoS attacks, Mirai attacks, MITM attacks, DNS attacks, reconnaissance attacks, vulnerability scans, brute force attacks, benign traffic, and other attacks. The total number of rows in the datasets is more than 1 million, where there are 820,525 samples of DDoS attacks, 181,481 samples of DoS attacks, 59,233 numbers of Mirai attacks, 7,019 samples of MITM attacks, 4,034 of DNS spoofing, 7,136 samples of recon attacks, 809 vulnerability scans, 3,244 brute force attacks, and 538 samples of other attacks. The dataset consists of 46 features. 



\subsection{Data Preprocessing}
Raw data is converted into a suitable format (raw input) so that it can be easily fed into the model (or pipeline). It is the first and most crucial step you take while building a machine-learning model. Some common problems in real-world data are noise, missing values, and null values, as well as issues such as the format of the velocity not being suitable for ML models. Data preparation procedures, which involve cleaning and organizing our data to feed it into the model, will, in some cases, help achieve better accuracy levels and speed up some occasions while building machine learning models. In the event that the dataset has missing data, it can seriously wreck our machine-learning model. Therefore, it is essential to handle the missing values in the dataset and complete those null values. Check for null and missing values in the dataset.
\subsection{Proposed Algorithms}
Network anomaly detection systems are the backbone of network security. Apart from identifying abnormalities, such systems for network anomaly detection constantly check and evaluate the information occurring within a network. With a publicly available dataset, the researchers checked five machine learning algorithms for abnormality identification. Here are the details:
\begin{itemize}
    \item K-Nearest Neighbor
    \item Gradient Boosting
    \item Decision Tree
    \item AdaBoost
    \item Random Forest
\end{itemize}

\subsection{Model Description}
The first algorithm used was the Random Forest Classifier. It consists of many independent trees, or trees of choice in RF, that are trained with training sample data in the end. The results from each of these trees are then put through a voting process to produce estimates. An RF classifier is used to calculate the final result based on the majority of votes. We used GridSearchCV for hyperparameter tuning. The tuned parameters are \enquote{criterion}: \enquote{gini}, \enquote{max\_depth}: 8, \enquote{max\_features}: \enquote{sqrt}, and \enquote{n\_estimators}: 200. Along with that, five-fold cross-validation was applied to help in mitigating bias and variance issues and to ensure consistent model performance on unseen data.

The second model put into practice was the Decision Tree Classifier. The method is widely applied in machine learning to solve problems related to regression and classification. The internal node and the leaf node are the two types of nodes that a root node produces. Because they have several branches, internal nodes are known as decision-makers, whereas leaf nodes are known for producing results because they do not have any more branches. We used GridSearchCV for hyperparameter tuning. The tested parameters are \enquote{criterion}: \enquote{entropy}, \enquote{max\_depth}: 30, \enquote{min\_samples\_leaf}: 5, \enquote{min\_samples\_split}: 10, and \enquote{max\_features}: \enquote{sqrt}. Along with that, five-fold cross-validation was applied to help in mitigating bias and variance issues and to ensure consistent model performance on unseen data.

The K-Nearest Neighbor approach classifies new data points according to how similar they are to preexisting data points, saving all available data. This suggests that new data may be easily classified into an appropriate category by applying the KNN algorithm. We used GridSearchCV for hyperparameter tuning. The tuned parameters are \enquote{n\_neighbors}: 5, \enquote{weights}: \enquote{distance}, \enquote{metric}: \enquote{manhattan}, and \enquote{p}: 1. Along with that, five-fold cross-validation was applied to help in mitigating bias and variance issues and to ensure consistent model performance on unseen data.

The gradient boosting algorithm (GBA) is one of the most effective machine learning techniques. Using this method, each predictor aims to minimize errors in order to outperform its prior performance. But the interesting idea behind gradient boosting is that it fits a new model to the regression that the previous predictor created instead of fitting a classifier to the data at each step~\cite{islam2025artificial}. We used GridSearchCV for hyperparameter tuning. The tested parameters are \enquote{learning\_rate}: 0.01, \enquote{max\_depth}: 4, \enquote{n\_estimators}: 500, and \enquote{subsample}: 0.8. Along with that, five-fold cross-validation was applied to help in mitigating bias and variance issues and to ensure consistent model performance on unseen data.

In machine learning, the AdaBoost algorithm is an ensemble method that uses boosting techniques. In this process, weights are reallocated to each instance, with greater weights assigned to incorrectly identified instances. We used GridSearchCV for hyperparameter tuning. The tuned parameters are \enquote{algorithm}: \enquote{SAMME.R}, \enquote{learning\_rate}: 0.1, and \enquote{n\_estimators}: 100. Along with that, five-fold cross-validation was applied to help in mitigating bias and variance issues and to ensure consistent model performance on unseen data.

\subsection{Workflow Diagram}
\begin{figure}[h]
    \centering
    \includegraphics[width=0.45\textwidth]{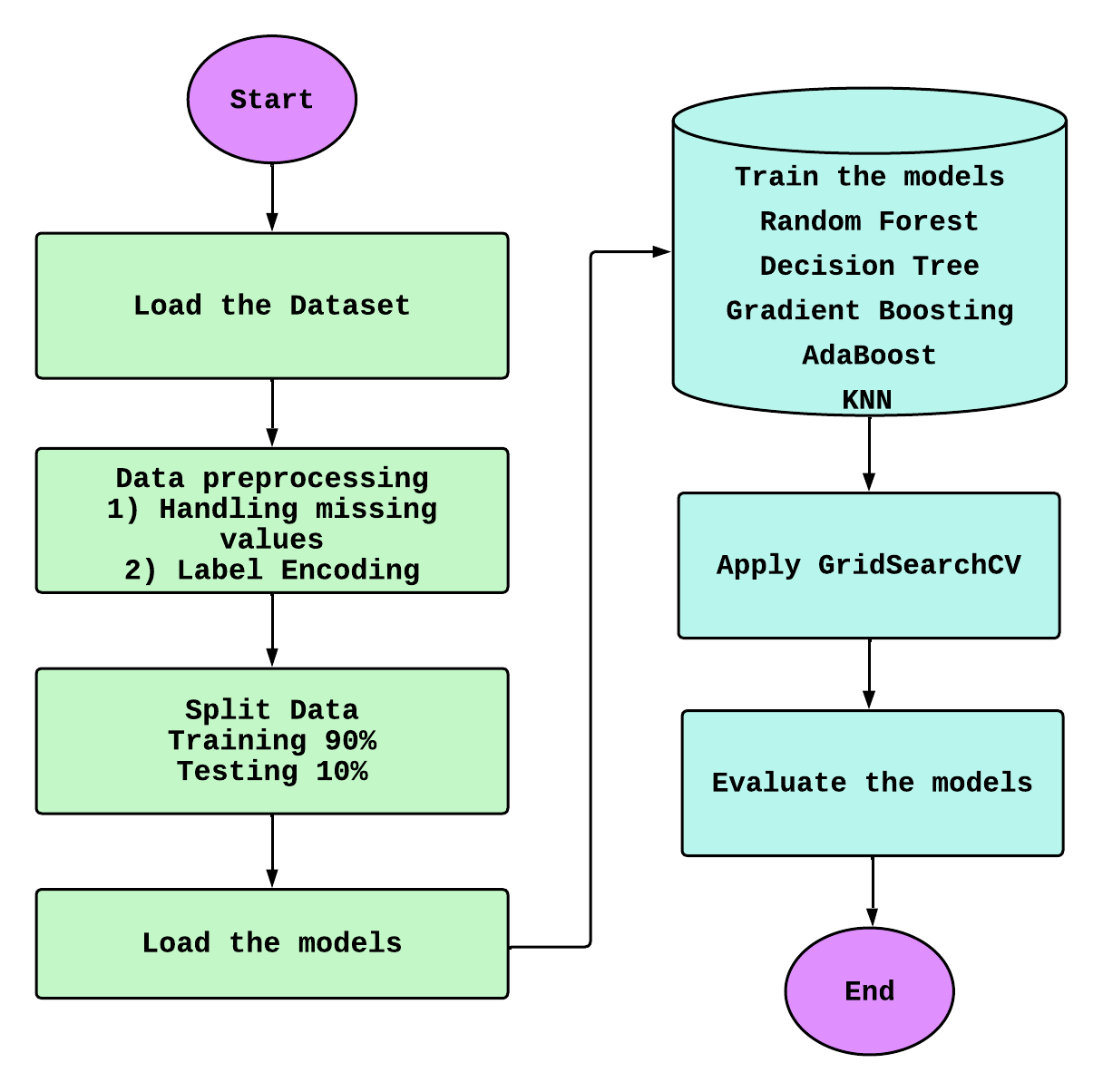}
    \caption {Workflow diagram of the system}
    \label{fig:2}
\end{figure}

The workflow diagram for this study is shown in Fig.~\ref{fig:2}. First, the dataset is preprocessed for missing data, handling of imbalanced data, and encoding on the level. Then, the data was divided in an 80-20 ratio. 80\% were used to model training, and 20\% were used for the models' testing. We execute and test the models like Random Forest (RF), Decision Tree (DT), K-Nearest Neighbor (KNN), Gradient Boosting, and AdaBoost. For detecting array models, the accuracy score obtained from these models will be used in the comparative analysis.

\subsection{Testing Methodology}
This subsection describes the test for the models in this study. The equations of the testing metrics are given below~\cite{islam2024deep,mahmud2024advance}:
\begin{equation}
\label{eq:precision}
\text{Precision} = \frac{TP}{TP + FP}
\end{equation}

\begin{equation}
\label{eq:recall}
\text{Recall} = \frac{TP}{TP + FN}
\end{equation}

\begin{equation}
\label{eq:F1}
F1-score= 2 \times \frac{\text{Precision} \times \text{Recall}}{\text{Precision} + \text{Recall}}
\end{equation}

\begin{equation}
\label{eq:accuracy}
\text{Accuracy} = \frac{TP + TN}{TP + TN + FP + FN}
\end{equation}

\section{Result Analysis}
The models were evaluated on the test set in terms of the precision, recall, F1 score, accuracy, confusion matrix, and ROC curves. The best two and the worst model's performance metrics and ROC curves are represented in this section with only the best model's confusion matrix.

\subsection{Decision Tree}

\begin{figure}[h]
    \centering
    \includegraphics[width=0.45\textwidth]{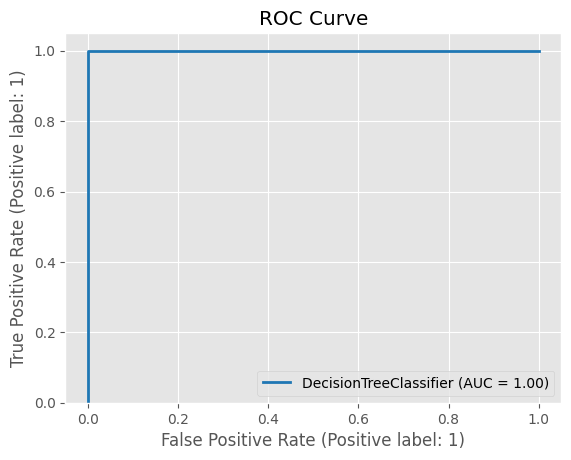}
    \caption{ROC curve of Decision Tree Classifier}
    \label{fig:10}
\end{figure}

The ROC curve of the Decision Tree classifier is shown in Fig. ~\ref{fig:10}. This classifier performed well with an AUC of 1.00. An AUC score of 1.0 means great distinguishing ability of the model, which can separate the two classes perfectly with almost no false positives and no false negatives. An ideal classifier has the max true positive rate and min false positive rate; hence, the ROC curve must pass through the left-hand corner of the plot. This should indicate overfitting or an extremely well-separated dataset.

\begin{figure*}[htbp]
\centering
\includegraphics[width=\textwidth]{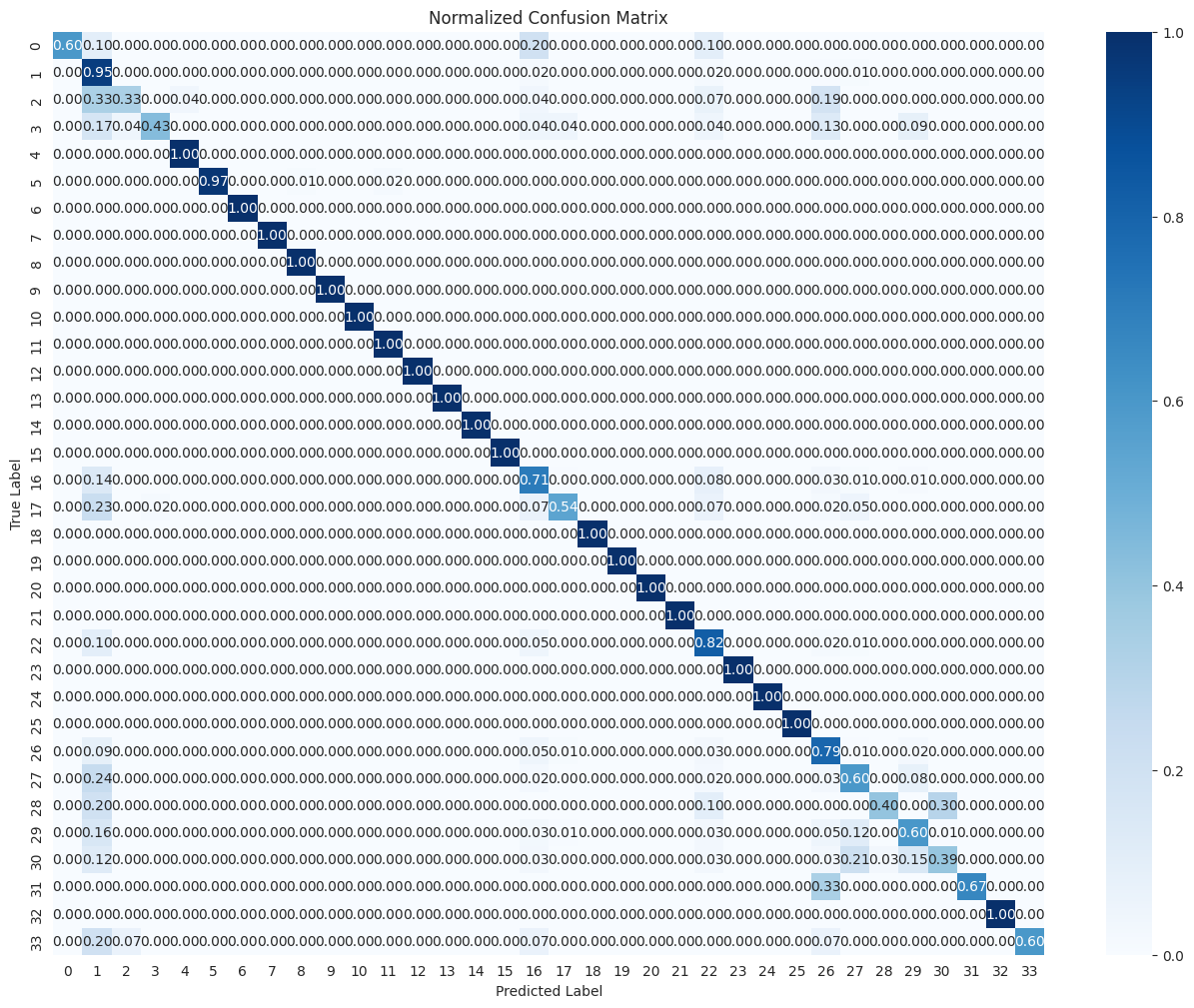}
\caption{Normalized Confusion Matrix of the multi-class attack classification by Decision Tree classifier model.}
\label{fig:CM_DT}
\end{figure*}

Fig. ~\ref{fig:CM_DT} presents the confusion matrix of the decision tree model. A high accuracy for a few classes (for example, 1, 4, and 33) can be found in certain diagonal entries, which represent correct predictions (the deeper the shadow, the more true predictions). However, the lighter colors in the matrix also indicate relatively low predictive accuracies between several classes, most visibly between 2 and 3 classes and also classes 30 and 33. This very detailed visualization helps to understand specific aspects of the model, which can be further fine-tuned or trained.

\subsection{Random Forest}

\begin{figure}[h]
    \centering
    \includegraphics[width=0.45\textwidth]{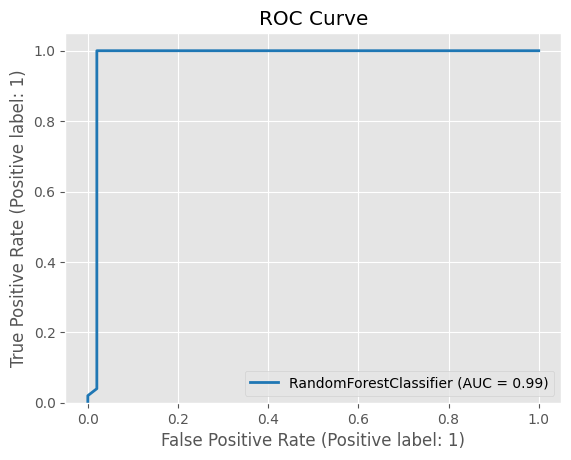}
    \caption{ROC curve of Random Forest Classifier}
    \label{fig:8}
\end{figure}

In Fig.~\ref{fig:8}, The ROC curve of the Random Forest classifier is shown with an AUC value of 0.99. The AUC value indicates that the model is able to discriminate well between the classes, and well below there are few exceptions, with a high rate of true positives. The reason why the curve goes quite close to the top left corner of the plot is that it effectively finds the positive class while keeping the false positives low, which indicates good performance of the classifier in practice settings.

\subsection{K-Nearest Neighbor}
\begin{figure}[h]
    \centering
    \includegraphics[width=0.45\textwidth]{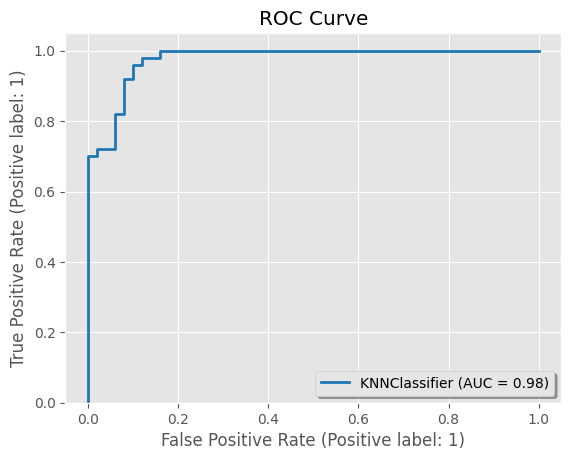}
    \caption{ROC cruve of K-Nearest Neighbor classifier}
    \label{fig:13}
\end{figure}

Fig. \ref{fig:13} shows the ROC curve for a KNN classifier, which has an AUC value of 0.98. A model achieving a high true positive rate at a low false positive rate means that the curves are close to the x-axis, indicating the model's ability to separate the classes, making the AUC very high. Because the KNN classifier has multiple threshold values, it is demonstrated in the step shape of the ROC curve, where we can observe the behavior of the classifier at different levels of operation.

\subsection{Model Evaluation}
Performance metrics comparison of different machine learning models as shown in table \ref{tab:1}. We will evaluate metrics such as precision, recall, F1 score, and accuracy (in percentage). The Decision Tree algorithm outperforms any other algorithm, with an F1 Score of 0.996 and an accuracy of 99.56\%; it shows almost perfect classification performance. The random forest and the gradient boosting model are also doing quite well, with precision, recall, and F1 scores all over 0.98 and accuracy rates just below that of the decision tree. In metrics like these, AdaBoost and K-Nearest Neighbor models are effective, but then their performance is definitely lower.

\begin{table}[htbp]
 \scriptsize
\caption{Performance Metrics Comparison}
\centering
\begin{tabular}{|c|c|c|c|c|}
\hline
\textbf{Model} & \textbf{Precision} & \textbf{Recall}&\textbf{F1 Score} & \textbf{Accuracy (\%)}\\
\hline
Random Forest & 0.981 & 0.982 & 0.982 & 98.22\\
\hline
Decision Free & 0.997 & 0.995& 0.996 & 99.56\\
\hline
Gradient Boosting & 0.981 & 0.971 & 0.982 & 98.19 \\
\hline
AdaBoost & 0.972 & 0.945 & 0.966 & 96.26\\
\hline
K-Nearest Neighbor &  0.963 & 0.955 & 0.962 & 96.11\\
\hline
\end{tabular}
\label{tab:1}
\end{table}

\subsection{Result Comparison}
The comparison of the models with those previously studied is shown in Table~\ref{tab:2}. It is evident from the table that the Decision Tree model overpowers all the frameworks. 

\begin{table}[htbp]
\caption{Result Comparison}
\centering
\begin{tabular}{|c|c|c|c|c|}
\hline
\textbf{Study}  & \textbf{Best Model} & \textbf{Accuracy (\%)} \\
\hline
This paper & \textbf{Random Forest} & \textbf{99.56} \\
\hline
\cite{10928532} & Random Forest & 99.39\\
\hline
\cite{mohy-eddine2022ensemble} & Random Forest & 99.30 \\
\hline
\cite{rose2021intrusion} & Ensemble & 98.35 \\
\hline
\cite{musleh2023intrusion} & KNN & 98.30 \\
\hline
\end{tabular}
\label{tab:2}
\end{table}

\section{Conclusion and Future Work}
The machine learning models employed are particularly powerful in their application towards IoT security, with the Decision Tree and Random Forest classifiers stemming mainly from the diverse range of literature investigated in this research paper. Experimental results on the CICIoT 2023 dataset show that our methods outperform the existing methods to detect and classify multiple types of attacks in IoT networks. The Decision Tree algorithm showed unprecedented performance at an accuracy of 99.56\% and an F1 score of 99.62\%, which verifies its applicability as a reliable algorithm for IoT security applications. Another model that performed quite well was Random Forest, where accuracy was about 98.22\% and the F1 score was 98.24\%. Such classifiers are ideally suited for the multifaceted, dynamic nature of Internet of Things environments. In addition, the ML-based approach has been shown to be effective in decreasing the variance and bias, leading to better prediction performance. However, opportunities for improvement are vast, especially with the application of advanced machine learning techniques to improve IoT security systems. Future work may analyze a greater variety of ensemble methods like boosting or stacking to enhance model robustness and precision. Moreover, the integration of real-time detection systems capable of promptly identifying and neutralizing threats as they manifest would significantly enhance response procedures and the overall effectiveness of security measures. And adaptive learning models that help counter evolving security threats will mean that defenses are not only effective but continually improve over time. Additionally, it would be essential to develop energy-efficient algorithms that can achieve high performance without exhausting device resources, given the resource constraints on many IoT devices. Lastly, this would have the potential to strengthen the generalizability and effectiveness of security solutions since it would offer broadly applicable protection across the ecosystem of connected technologies by validating security for a wider class of IoT devices from diverse vendors.

\bibliographystyle{IEEEtran}
\bibliography{IEEE.bib}

\end{document}